\documentclass[sigconf]{acmart}

\usepackage{svg}
\usepackage{multirow}
\usepackage{pifont}

\AtBeginDocument{%
  }

\copyrightyear{2025}
\acmYear{2025}
\setcopyright{acmlicensed}\acmConference[WWW Companion '25]{Companion Proceedings of the ACM Web Conference 2025}{April 28-May 2, 2025}{Sydney, NSW, Australia}
\acmBooktitle{Companion Proceedings of the ACM Web Conference 2025 (WWW Companion '25), April 28-May 2, 2025, Sydney, NSW, Australia}
\acmDOI{10.1145/3701716.3715252}
\acmISBN{979-8-4007-1331-6/2025/04}

\def\b#1{\mathbf{#1}}
\def\c#1{\mathcal{#1}}
\def\bb#1{\mathbb{#1}}

\def\t#1{\text{#1}}
\def\({{\big (}}
\def\){\big )}

\makeatletter
\newcommand*{\bigcdot}{}
\DeclareRobustCommand*{\bigcdot}{%
  \mathbin{\mathpalette\bigcdot@{}}%
}
\newcommand*{\bigcdot@scalefactor}{.5}
\newcommand*{\bigcdot@widthfactor}{1.15}
\newcommand*{\bigcdot@}[2]{%
  \sbox0{$#1\vcenter{}$}
  \sbox2{$#1\cdot\m@th$}%
  \hbox to \bigcdot@widthfactor\wd2{%
    \hfil
    \raise\ht0\hbox{%
      \scalebox{\bigcdot@scalefactor}{%
        \lower\ht0\hbox{$#1\bullet\m@th$}%
      }%
    }%
    \hfil
  }%
}
\makeatother

\makeatletter

\def\env@sqcases{
  \let\@ifnextchar\new@ifnextchar
  \left\lbrack
  \def\arraystretch{1.2}%
  \array{@{}l@{\quad}l@{}}%
}
\makeatother

\makeatletter
\newcommand\footnoteref[1]{\protected@xdef\@thefnmark{\ref{#1}}\@footnotemark}
\makeatother

\begin{document}


\title{On the Practice of Deep Hierarchical Ensemble Network for Ad
Conversion Rate Prediction}

\author{Jinfeng Zhuang}
\affiliation{%
  \institution{Pinterest Inc.}
  \streetaddress{651 Brannan St}
  \city{San Francisco}
  \state{California}
  \country{USA}
}
\email{jzhuangl@pinterest.com}

\author{Yinrui Li}
\affiliation{%
  \institution{Pinterest Inc.}
  \streetaddress{651 Brannan St}
  \city{San Francisco}
  \state{California}
  \country{USA}
}
\email{yinruili@pinterest.com}

\author{Runze Su}
\affiliation{%
  \institution{Pinterest Inc.}
  \streetaddress{651 Brannan St}
  \city{San Francisco}
  \state{California}
  \country{USA}
}
\email{runzesu@pinterest.com}

\author{Ke Xu}
\affiliation{%
  \institution{Pinterest Inc.}
  \streetaddress{651 Brannan St}
  \city{San Francisco}
  \state{California}
  \country{USA}
}
\email{kxu@pinterest.com}

\author{Zhixuan Shao}
\affiliation{%
  \institution{Pinterest Inc.}
  \streetaddress{651 Brannan St}
  \city{San Francisco}
  \state{California}
  \country{USA}
}
\email{zshao@pinterest.com}

\author{Kungang Li}
\affiliation{%
  \institution{Pinterest Inc.}
  \streetaddress{651 Brannan St}
  \city{San Francisco}
  \state{California}
  \country{USA}
}
\email{kungangli@pinterest.com}

\author{Ling Leng}
\affiliation{%
  \institution{Pinterest Inc.}
  \streetaddress{651 Brannan St}
  \city{San Francisco}
  \state{California}
  \country{USA}
}
\email{lleng@pinterest.com}

\author{Han Sun}
\affiliation{%
  \institution{Pinterest Inc.}
  \streetaddress{651 Brannan St}
  \city{San Francisco}
  \state{California}
  \country{USA}
}
\email{hsun@pinterest.com}

\author{Meng Qi}
\affiliation{%
  \institution{Pinterest Inc.}
  \streetaddress{651 Brannan St}
  \city{San Francisco}
  \state{California}
  \country{USA}
}
\email{mengqi@pinterest.com}

\author{Yixiong Meng}
\affiliation{%
  \institution{Pinterest Inc.}
  \streetaddress{651 Brannan St}
  \city{San Francisco}
  \state{California}
  \country{USA}
}
\email{ymeng@pinterest.com}

\author{Yang Tang}
\affiliation{%
  \institution{Pinterest Inc.}
  \streetaddress{651 Brannan St}
  \city{San Francisco}
  \state{California}
  \country{USA}
}
\email{ytang@pinterest.com}

\author{Qifei Shen}
\affiliation{%
  \institution{Pinterest Inc.}
  \streetaddress{651 Brannan St}
  \city{San Francisco}
  \state{California}
  \country{USA}
}
\email{qshen@pinterest.com}

\author{Zhifang Liu}
\affiliation{%
  \institution{Pinterest Inc.}
  \streetaddress{651 Brannan St}
  \city{San Francisco}
  \state{California}
  \country{USA}
}
\email{zhifangliu@pinterest.com}

\author{Aayush Mudgal}
\affiliation{%
  \institution{Pinterest Inc.}
  \streetaddress{651 Brannan St}
  \city{San Francisco}
  \state{California}
  \country{USA}
}
\email{amudgal@pinterest.com}

\author{Caleb Lu}
\affiliation{%
  \institution{Pinterest Inc.}
  \streetaddress{651 Brannan St}
  \city{San Francisco}
  \state{California}
  \country{USA}
}
\email{klu@pinterest.com}

\author{Jie Liu}
\affiliation{%
  \institution{Pinterest Inc.}
  \streetaddress{651 Brannan St}
  \city{San Francisco}
  \state{California}
  \country{USA}
}
\email{jieliu@pinterest.com}

\author{Hongda Shen}
\affiliation{%
  \institution{Pinterest Inc.}
  \streetaddress{651 Brannan St}
  \city{San Francisco}
  \state{California}
  \country{USA}
}
\email{hshen@pinterest.com}
\renewcommand{\shortauthors}{Jinfeng Zhuang et al.}

\begin{abstract}
The predictions of click through rate (CTR) and conversion rate (CVR) play a crucial role in the success of ad-recommendation systems. A Deep Hierarchical Ensemble Network (DHEN) has been proposed to integrate multiple feature crossing modules and has achieved great success in CTR prediction. However, its performance for CVR prediction is unclear in the conversion ads setting, where an ad bids for the probability of a user's off-site actions on a third party website or app, including purchase, add to cart, sign up, etc. Because of the ensemble nature, the degree of freedom of DHEN is essentially high, for example, 1) What feature-crossing modules (MLP, DCN, Transformer, to name a few) should be included in DHEN? 2) How deep and wide should DHEN be to achieve the best trade-off between efficiency and efficacy? 3) What hyper-parameters to choose in each feature-crossing module? Orthogonal to the model architecture, the input personalization features also significantly impact model performance with a high degree of freedom. It is an important and interesting problem in the advertising industry on how to make DHEN work effectively with a diverse collection of personalization features for web-scale CVR prediction. In this paper, we attack this problem and present our contributions biased to the applied data science side, including: 

First, we propose a multitask learning framework with DHEN as the single backbone model architecture to predict all CVR tasks, with a detailed study on how to make DHEN work effectively in practice; Second, we build both on-site real-time user behavior sequences and off-site conversion event sequences for CVR prediction purposes, and conduct ablation study on its importance; Last but not least, we propose a self-supervised auxiliary loss to predict future actions in the input sequence, to help resolve the label sparseness issue in CVR prediction.


Our method achieves state-of-the-art performance compared to previous single feature crossing modules with pre-trained user personalization features. It has been deployed in our Pinterest conversion ad recommendation system and has significantly boosted both user value and advertiser value by connecting online inspiration to real-world actions.

\end{abstract}



\begin{teaserfigure}
  \includegraphics[width=\textwidth]{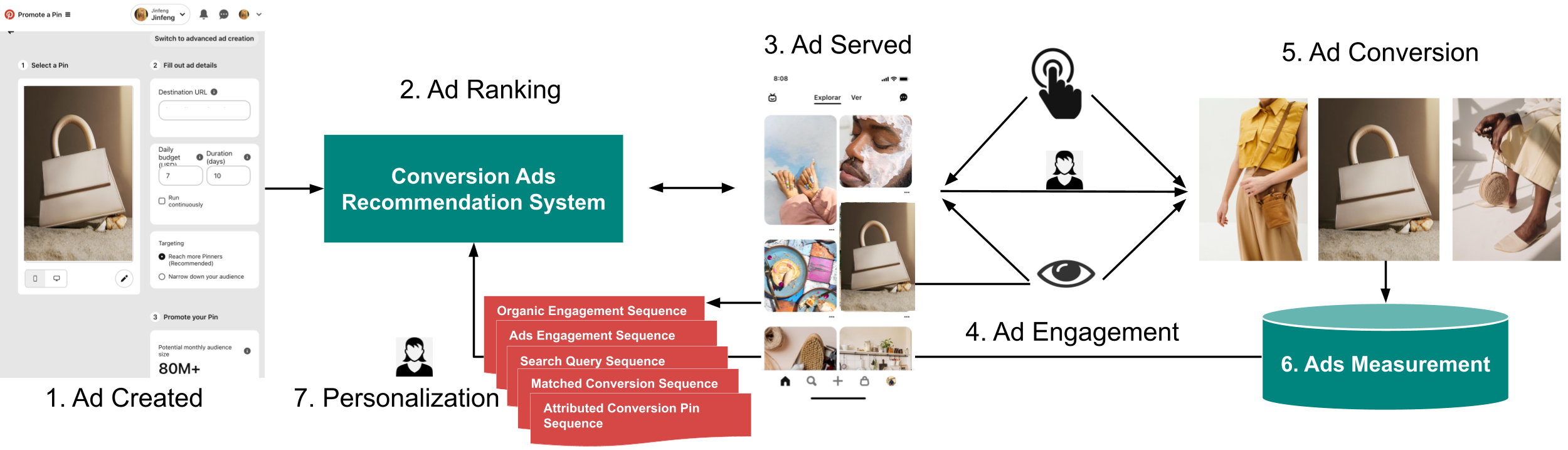}
  \caption{Illustration of the workflow of conversion ad recommendation with use behavior sequence modeling.}
  \label{fig:teaser}
\end{teaserfigure}



\maketitle

\section{Introduction}\label{sec:introduction}

With more than 500 million monthly active users, Pinterest has become an important visual inspiration platform for people to search, save, and shop the best ideas in the world for all of life's moments. The native advertising format of Pinterest presents merchants' advertising content to a user and helps a user decide what to purchase off-site, often with an image or video, title, and well-designed description.


A particular ad type called conversion ad, also known as Optimized Cost Per Mille (oCPM) ad in the advertising industry, aims to optimize the rate of user's off-site conversion after impressed on the ad, including checkout, add to cart, sign up, etc. It is a key pillar of an online monetization system as well as a key component to bridge the gap between users' online inspiration and their off-site realization. 

Predicting the conversion rate (CVR) of an oCPM ad accurately and ranking the conversion ads on top of it can dominate the quality and efficiency of the advertising delivery funnel. Therefore, CVR prediction has been crucial, in general, for the success of the digital advertising business. Besides the importance, it is also a technically interesting problem in a social media's ecosystem, given that both the users' behavior patterns and the data format from the content side are diverse. It takes cutting-edge machine learning techniques to exploit the value in these rich input data, such that the user's value, merchant's value, and platform's business value are jointly maximized.

CVR prediction is different from onsite engagement prediction in nature, e.g., click through rate (CTR) prediction, because: 1) the volume of labels is sparser; 2) the delay of collecting labels is longer; 3) the label itself is more noisy. Our high-level recipe for these challenges is better measurement of off-site conversions to improve label quality, and better user understanding to maximize the modeling power on given labels. We limit the scope of this paper to the latter: we optimize serving every user ad content that is highly personalized to the user's interests, tastes, goals and intent given a fixed mechanism of collecting conversion labels\footnote{\label{note_privacy}Pinterest serves personalized ads and uses off-site conversion data for users only when allowed pursuant to user privacy choices and applicable laws.}.

However, personalized CVR prediction is a challenging problem~\cite{li2024privacy}. There is no golden standard for a one-fit-all solution to featurize the diverse data format, and there are many existing feature crossing modules that can handle a given collection of feature vectors. The degree of freedom is essentially high. In addition, how to squeeze the power of the sparse conversion labels has not been explored much in the past.


Previously, from the personalization perspective, a pre-trained user embedding that can be plugged into retrieval and ranking models has achieved great success at Pinterest~\cite{PalEZZRL20,PanchaZLR22,XuZR22,ZhuangL19,YingHCEHL18,deWetO19, hsu2024taming}. In particular, the PinnerFormer model~\cite{PanchaZLR22,XuZR22} that summarizes a user engagement sequence into an embedding vector in a Euclidean space has produced a step function change in home feed recommendation. However, PinnerSAGE or PinnerFormer is optimized for onsite engagement only, which is not optimized for the conversion probability over impression of the ad. Their training data do not include off-site conversion data\footnoteref{note_privacy}.

From the model architecture perspective, many Deep Neural Networks (DNN) models have been proposed to handle feature crossing, biased to CTR prediction in the ad industry, e.g., Multilayer Perceptron~\cite{rosenblatt1958perceptron}, Transformer~\cite{VaswaniSPUJGKP17}, DeepFM~\cite{GuoTYLH17}, DCN~\cite{WangFFW17,WangSCJLHC21, wang2023towards}, FibiNet~\cite{huang2019fibinet, zhang2023fibinet++},  MaskNet~\cite{abs-2102-07619}, DIN~\cite{ZhouZSFZMYJLG18,FengLSWSZY19,abs-1809-03672}, User Behavior Sequence Modeling~\cite{abs-1905-09248,abs-2006-05639,abs-2205-10249,ChangZFZGLHLNSG23}, just to name a few. In order to take advantage of the strengths of different models, a Deep Hierarchical Ensemble Network (DHEN)~\cite{abs-2203-11014, zhang2024wukong} has been proposed to incorporate different feature crossing modules as a "cocktail" solution. It has been applied in Meta's CTR prediction successfully and achieved great business success. It is also technically interesting what modules and configurations make DHEN work in applications.



With a focus on optimizing the CVR prediction with DHEN, we present the most important components in the practice of developing a personalized web-scale CVR prediction model. Our contributions and key value proposals include:
\begin{itemize}
    \item Build a unified multitask model for all off-site conversion actions with DHEN as the backbone architecture. We conduct detailed analysis and ablation study on what modules are best working;
    \item Model user action sequence end-to-end in the CVR prediction task, and prove that it adds significant values in addition to pre-trained embeddings; 
    \item Introduce a self-supervised future action prediction loss and it can benefit supervised tasks like CVR prediction.
\end{itemize}

We deployed the DHEN model in Pinterest's ad recommendation system in 2024 and boosted top-line business metrics like Cost-Per-Acquisition (CPA) significantly. The long-term gain in engineering is also significant: we build a framework where we can keep iterating new feature-crossing modules with a built-in self-supervised learning mechanism to address the label sparseness problem. The feature engineering paradigm is shifted to 1) add more types of sequence data to the input; 2) enrich the information per item in the sequence, from designing handcrafted and pre-trained features. 

We elaborate the key design decisions in section~\ref{sec:design}, the detailed model architecture in Section~\ref{sec:model}. Both off-line evaluation and online results are presented in Section~\ref{sec:exp}.

\section{Problem Setup and High-Level Design}\label{sec:design}

In this section, we formulate the CVR prediction problem and elaborate the key design decisions.

\subsection{Problem Setup} 

We start with a corpus of "pins" $\c{P} = \big\{P_1, P_2, ..., P_N \big\}$, where $N$ is
a large-scale value on the order of billions, and a set of users $\c{U} = \big\{U_1, U_2, ..., U_M\big\}$,
where $M$ is a large-scale value on the order of millions. Each pin here is classified into \textit{organic content} and \textit{advertising content}. We also have access to the sequence of both onsite actions (click, long click, download, hide, etc.) and off-site conversion actions (checkout, add to cart, sign up, etc.) where allowed pursuant to user privacy choices.

Conversion rate prediction is essentially the modeling of the possibility that a user would convert for the advertiser, that is, take expected off-site actions such as checkout, when an ad content bid by this advertiser is presented in front of the user: $f: \c{U}\times\c{P}\rightarrow \bb{R}^+$. This is usually formulated by the click-based conditional probability:
\[
\bb P(\t{conv}; U, P) = \bb P(\t{ctr}) * \bb P\big(\t{conv}\; |\; \t{ctr}\big),
\]
where $\bb P(\t{ctr})$ is the Click-Through Rate (CTR) and $\bb P(\t{conv}\; |\; \t{ctr})$ is the probability of $U$ converts after clicking $P$, respectively. It is also not uncommon for impression-based formulations to be used in industry; we omit this case without loss of generality.

\subsection{Multi-Head DNN with CTR Prediction}

As mentioned above, the CVR prediction problem naturally has orders of magnitude less labels than the CTR prediction. The top design choice is how to increase the number of labels. Our solution to this problem is to:
\begin{itemize}
    \item Merge all conversion tasks together and learn a unified model to predict them in a single forward inference, instead of building one model per task;
    \item Include CTR prediction head in the training process for CVR modeling purpose. This head brings much training labels and will not be used in online inference;
    \item Then the final supervised objective function would be a weighted loss defined over each head. We use binary cross-entropy (BCE) as the loss function for each head.  
\end{itemize}

Mixing all prediction tasks into a single Multi-task Learning (MTL) setting significantly increases the training data. The CTR head has proved very useful to regularize the parameter space and allows us to use a much larger model than CVR tasks only. We choose DHEN~\cite{abs-2203-11014} as the architecture of the backbone model, by virtue of the ensemble of different feature-crossing modules. Historically, our model evolved through different single feature crossing modules. DHEN has been shown to be capable of taking advantage of each feature-crossing module. Figure~\ref{fig:mtl-arch} presents the overall architecture we used.

\subsection{Hand Crafted Features VS Sequence based User Features} \label{sec:seq}

Feature engineering is an important pillar in the success of CVR prediction. Some representative examples include: 1) engagement counting features in a past rolling time window on a particularly entity, like ad campaign, advertiser, web domain, etc.; 2) predicted category features like interest or product category on a predefined taxonomy thesaurus; 3) pre-trained user embeddings and content embeddings.


We argue that for many types of user feature, similar to ~\cite{abs-2402-17152}, can be learned directly by a DNN end-to-end, if the crafted feature is extracted from the past user's behavior sequence. For example, the engagement count can simply be implemented as a sum-pooling module over a past time window. Predicted category features and pre-trained embedding features are also essentially derived from the user's behavior sequence. Therefore, we focus on exploring what types of sequences to pass to the model and what properties to exploit per item in the sequence. We will discuss the sequence modeling methods in Section~\ref{sec:model}.

\textbf{Organic sequence VS ads specific sequence.} It has been reported that the user onsite engagement sequence is a very strong feature to build user representations~\cite{PanchaZLR22,PalEZZRL20}. When we zoom into the CVR prediction problem, the sequence data can be extended to the 3 types of sequences, respectively:

\begin{itemize}
\item $S_{\t{search}}$: Search Query Sequence, consisting of user's past search queries at Pinterest;
\item $S_{\t{org}}$: Organic Content Engagement Sequence, consisting of the content $P^O$ that are not ads generated by advertisers;
\item $S_{\t{ads}}$: Ad-content engagement sequence. Note that we do not distinguish the type of the ad product, i.e., both traffic ads and conversion ads are included in $S_{\t{ads}}$;
\end{itemize}

Both $S_{\t{org}}$ and $S_{\t{ads}}$ are a sequence of pins, which means that all the metadata and pre-trained signals are available on each item~\cite{HamiltonYL17,YingHCEHL18,ZhaiWTPR19}. For completeness, we define the input signals for each pin by a union of ID level signals.


\textbf{Onsite sequence VS off-site sequence.} oCPM ads rely on off-site conversion events to measure its performance. It involves \textit{User Match} between conversion and user, and \textit{Attribution}, to identify which ad presented to a user leads to their conversions. We define two off-site sequences that have stronger causality with CVR than onsite engagement sequences:

\begin{itemize}
\item $S_{\t{match}}$: Matched Conversion Sequence, where each item is a pair of user ID and advertiser ID, together with the off-site action type. It is not known which ad is leading to conversion yet;
\item $S_{\t{conv}}$: Attributed Conversion Sequence, consisting of the ad content in which each ad is attributed to an off-site conversion action.
\end{itemize}

Putting together, our input on the personalization features would be an ordered list of user behavior sequences:
\begin{equation}
    \c{S} = \big{\{} S_{\t{search}}, S_{\t{org}}, S_{\t{ads}}, S_{\t{match}}, S_{\t{conv}} \big{\}} \label{seqs}
\end{equation}

We tried merging all sequences into a single one ordered by timestamp, but we found that keeping them separate is better and provides more flexibility. The open question here is how to use the power encoded in $\c{S}$ to help predict the CVR.

\subsection{Pre-trained User Embedding VS End-to-end Sequence Modeling}

The pre-trained user embedding is very powerful as a general-purpose personalization feature at Pinterest~\cite{GrbovicC18,deWetO19,PalEZZRL20,PanchaZLR22}. The other view of personalization is to model the user representation end-to-end in a particular recommendation model. It has been proven that it is possible to achieve significant gains~\cite{XiaEPBWGJFZZ23} by modeling the sequence directly in the home feed recommendation. However, it is unknown how it can increase the value in the ad recommendation scenario. We present our sequence models in Section~\ref{sec:model} and verify its impact in the experiment section.

\subsection{Batch-mode VS Real-time Sequence} 

This comparison is based on the implementation side. There is no difference in the modeling, as both can be handled by an identical architecture. Sequence data are both storage consuming and computationally heavy. It usually takes a separate service to aggregate real-time user actions. In practice, we find that it is a good strategy to start with the offline batch-mode sequence feature. Once it is launched, the gain from the freshness of real-time sequence is well justified and always provides additional value. In this paper, our sequence data contain both batch mode and real-time data.

\section{Multitask Ensemble Network with Self-Supervised Auxiliary Loss}\label{sec:model}

In this section, we present the forward architecture for the CVR prediction model.

\subsection{Deep Hierarchical Ensemble Network}

\begin{figure}
\centering
\includegraphics[width=\columnwidth]{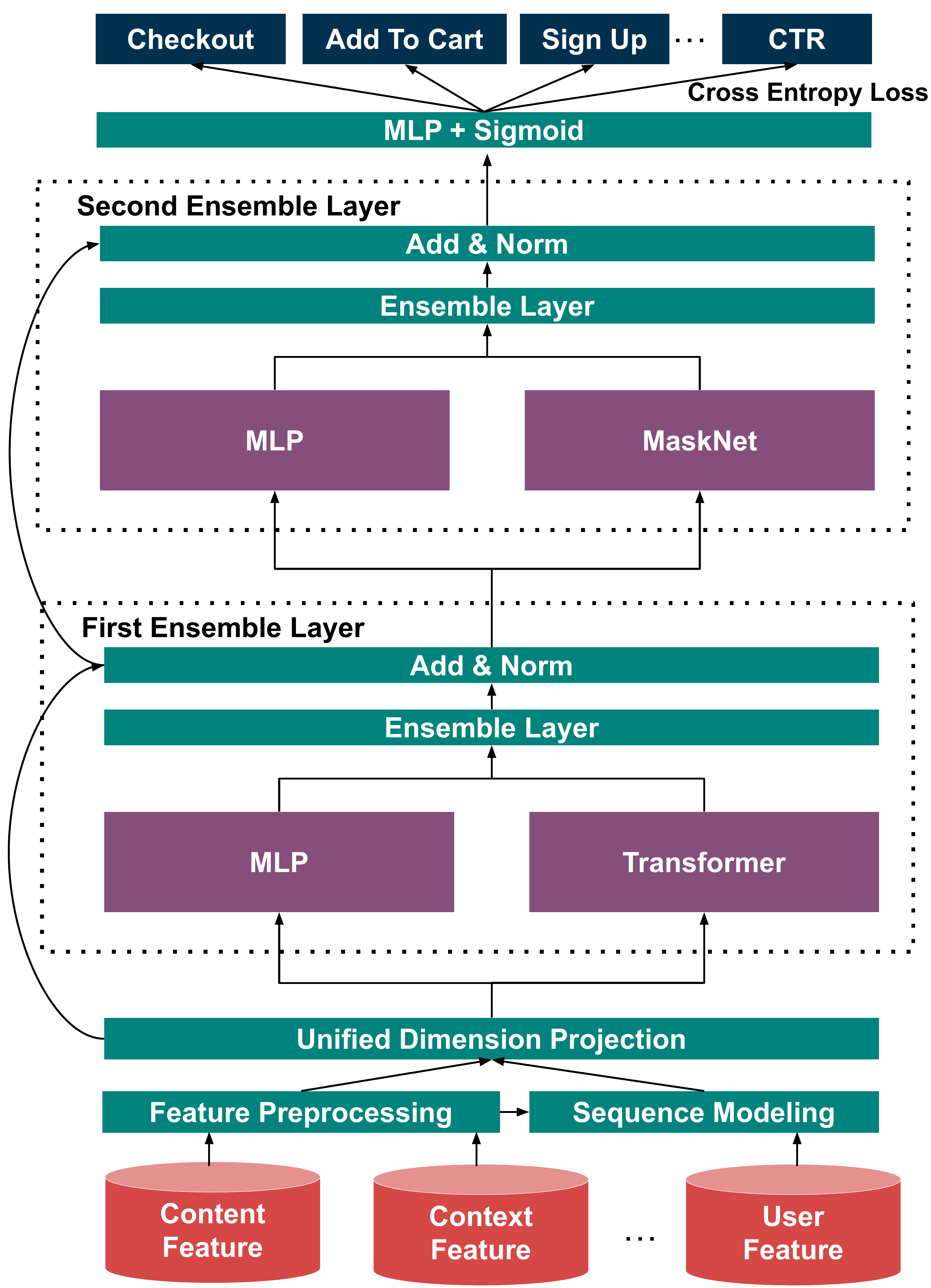}
\caption{The Multitask DHEN model architecture for CVR prediction. We found empirically the best trade-off is two layer of feature crossing, where the first layer is MLP + Transformer, and the second layer is MLP + MaskNet.}
\label{fig:mtl-arch}
\end{figure}

Figure~\ref{fig:mtl-arch} presents our backbone architecture following a DHEN~\cite{abs-2203-11014} style of feature interaction. The DHEN model is essentially a cocktail solution that assembles different feature interaction modules for the prediction of CTR. We found that it works well empirically for CVR prediction as well, with a careful selection of feature-crossing modules. We present some details on each component as follows:

\textbf{Feature Preprocessing.} We have four steps of preprocessing sequentially: 1) continuous feature normalization: we use min-max normalization to scale continuous features to $[0, 1]$; 2) batch normalization over pre-trained embedding features; 3) timestamp transformation: for the time at $i$, it will be $timestamp[i] = log(timestamp[i] - timestamp[i - 1] + 1.0)$; 4) categorical feature mapping to embedding vectors, which are learnable variables.

\textbf{Unified Dimension Projection.} The input tensors are projected to 3 dimension tensors $\mathbb{R}^{B\times L \times D}$, where $B$ is the batch size, $L$ is number of tokens, and $D$ is the dimension. The purpose of this layer is to make input generally compatible with the following feature-crossing modules.

\textbf{Feature-Crossing Modules.} We evaluated an extensive set of feature-crossing modules. Before DHEN as the ensemble model, we iterated through MLP~\cite{rosenblatt1958perceptron}, Transformer~\cite{VaswaniSPUJGKP17}, DeepFM~\cite{GuoTYLH17}, SENet~\cite{HuSS18}, DIN~\cite{ZhouZSFZMYJLG18}, DCN~\cite{WangFFW17}, DCN V2~\cite{WangSCJLHC21}, MaskNet~\cite{abs-2102-07619}, and HSTU~\cite{abs-2402-17152}. In practice, we found the ensemble approach with two layers of crossing modules achieves the best online performance within latency and infra cost budget, as presented in Figure~\ref{fig:mtl-arch}.

\textbf{Final MLP + Sigmoid.} It is a multilayer feedforward MLP plus activation function for each prediction head. Note that each head has its own MLP layers.

\textbf{Sequence Modeling.} The user's past behavior sequences defined in Section~\ref{sec:seq} are particularly important. We need to encode user sequence features before they can be fed to the projection layer. In our deployed system, we concatenate the vector of each feature per item as the representation of this item, then pass it to the Transformer Encoder to model the sequence.

The sequence feature is important for the model's performance because many hand-crafted features can be derived from a proper operator defined on the sequence. The adoption of Transformers is inspired by their success in language modeling. We further explore their power in solving the label sparseness issue.

\subsection{Self-Supervised Future Action Prediction}

\begin{figure}
\centering
\includegraphics[width=\columnwidth]{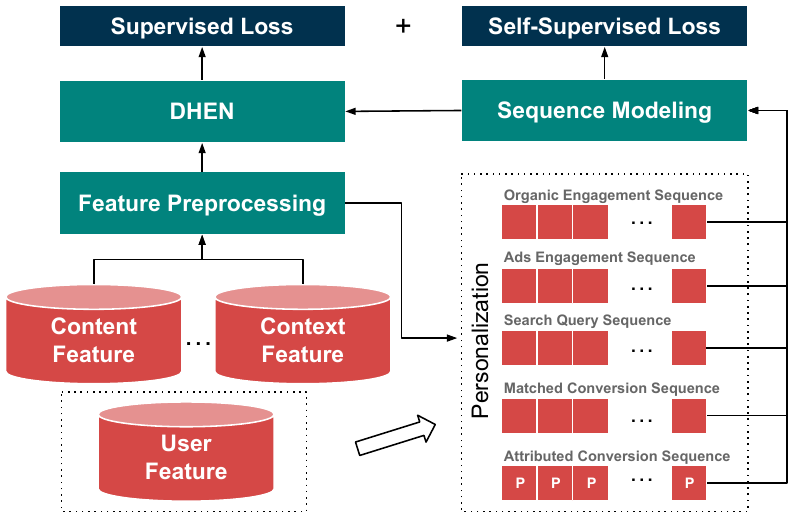}
\caption{The end-to-end sequence based personalization for conversion ads ranking. It introduces a self-supervised loss to predict future actions in input sequences, which helps to solve the label sparseness problem.}
\label{fig:seq-arch}
\vspace{-0.5cm}
\end{figure}

To exploit the interest of the user encoded in the user behavior sequences $\c{S}$, we propose a joint learning between CVR prediction and self-supervised future action prediction. We add a self-supervised loss item $J: \c{S} \rightarrow \mathbb{R}$: 
\[
L(U, P, Y) = \sum_{\t{heads}} \t{BCE\big(f(U, P), Y\big)} \;+\; \alpha * \sum_{S\in\c{S}} L\big( S \big),
\]
where $U$ denotes a user and $S$ is a behavior sequence of $U$, $P$ denotes an ad, $Y\in\{0, 1\}$ is a binary label for a specific conversion action between this pair of $(U, P)$, $f:\c{U}\times\c{P}\mapsto \bb{R}$ is the DNN model that predicts CVR, and $\t{BCE}$ is the binary cross-entropy function. The sum of $\t{BCE}$ is over different conversion action types, $\alpha$ is a hyperparameter controlling the importance of the sequence modeling loss.

Inspired by the idea of pre-trained user embedding~\cite{PanchaZLR22}, we enforce a self-supervised objective function, which defines a loss over the probability of generating target items in $S$ from the items happening before some timestamp $t_0$:
\begin{equation}
    L\big(S\big) = -\sum_{t_0 < t < N}\log \bb{P}\Big( P_{t} \;\Big|\;  P_{t^{\prime} < t} \Big), \label{eqn:gen}
\end{equation}

For web-scale applications, it is computationally prohibitive to model a probability distribution on $\bb{P}$. We take advantage of the $\t{INFO-NCE}$ loss~\cite{abs-1807-03748} to mock it. Let $\b{x}$ be the forward embedding of an ad $P$ from the encoder, then the probability in (\ref{eqn:gen}) is approximated by:

\begin{equation}
L \big( S \big) :=  -\sum_{t_0 < t <N} \log \frac{e^{ \b{x}^\top \b{x}_{t}} }{e^{\b{x}^\top \b{x}_{t}} + \sum_{P_{-} \in \c{N}_t}e^{ \b{x}^\top \b{x}_{-}}}, \label{eqn:info-nce}
\end{equation}
where $\c{N}_t$ is a collection of sampled negative pins for position $t$.

\subsection{ParetoNet Parameter Search}\label{sec:pareto}

Based on the above proposal, there is a large search space for hyperparameters, with limited computational resources for both training and online serving. We shall find the Pareto front with the performance and cost trade-off. 

\begin{figure}
\centering
\includegraphics[width=0.9\columnwidth]{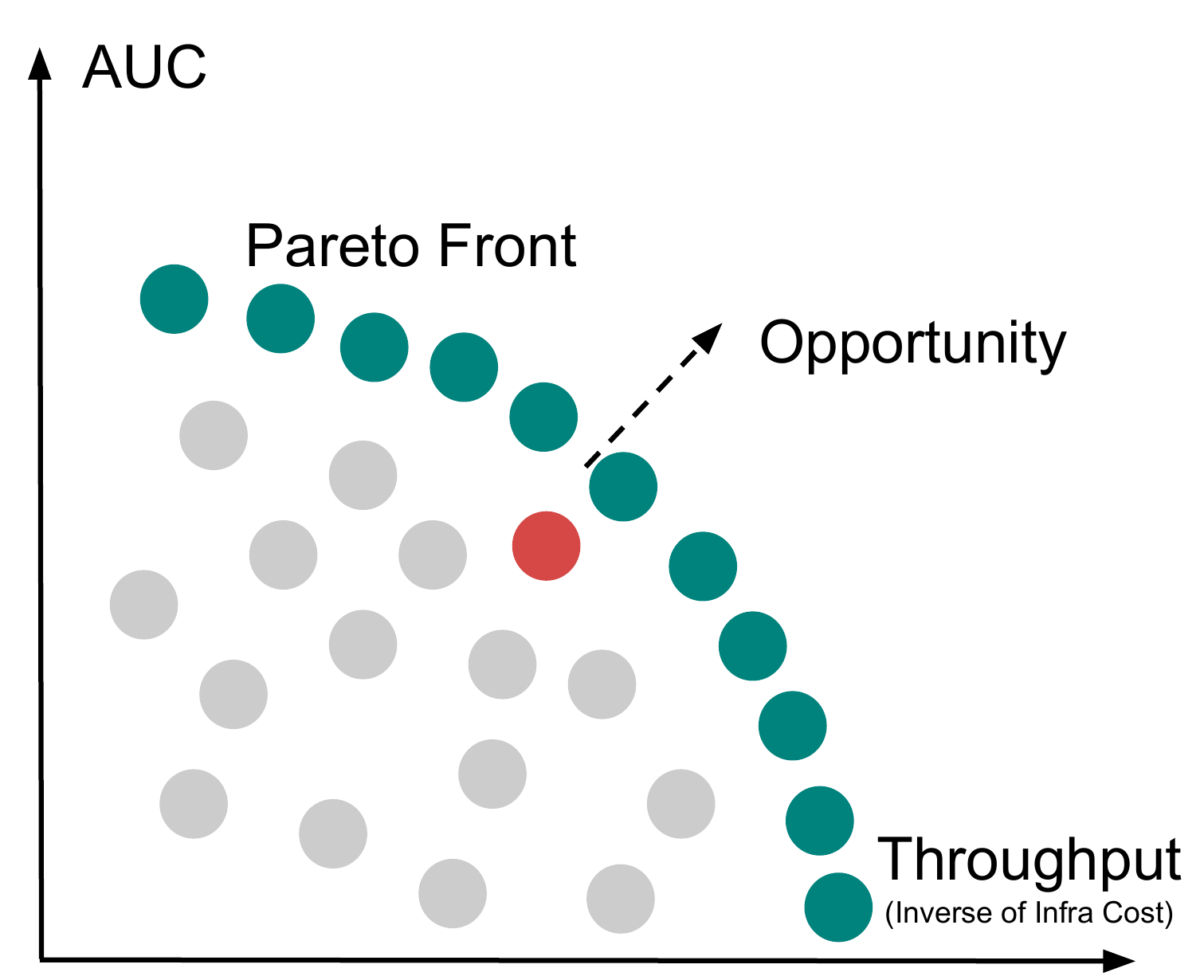}
\vspace{-0.4cm}
\caption{The Pareto front of model's performance with throughput / infra cost.}
\label{fig:pareto}
\vspace{-0.5cm}
\end{figure}

We use a simple, yet effective, neural architecture search (NAS) strategy for parameter search:
\begin{enumerate}
    \item Define and build the essential baseline minimum viable product model;
    \item Define the search space and dimensions to search for;
    \item Randomly sample from the search space and train the model to get AUC and Throughput. Use 60 day's data for training and 3 day's data for evaluation for efficiency purpose;
    \item Build a predictive model with search dimensions as input features and AUC/Throughput as prediction targets;
    \item Sample random points from the search space, and use the predictive model to estimate the AUC and Throughput.
    \item Identify the Pareto efficient candidates from Step 5 for optimal model building. 
\end{enumerate}

\section{Related Works}\label{sec:related-work}

For the model architecture part, our work is related to sequence modeling and feature crossing. For the learning paradigm side, it resides in the intersection of Contrastive Learning (CL), Self-supervised Learning (SSL), and Multi-task Learning (MTL).

\subsection{Feature Crossing in DNN}

Given the input features, a focus of DNN architecture design has been to cross the input features including embedding vectors to increase model's capacity. Crossing means multiplying two or more dimensions of input feature vectors. In general, any function that constructs an output dimension involving multiple input dimensions can be counted as a crossing. Some representative examples include ~\cite{rosenblatt1958perceptron,VaswaniSPUJGKP17,GuoTYLH17,HuSS18,ZhouZSFZMYJLG18,WangFFW17,WangSCJLHC21,abs-2102-07619,abs-2402-17152}, etc.

We evolved through a single feature crossing module in the history of CVR prediction and found that DHEN is capable of taking different advantages of different feature crossing modules, leading to the final implementation of our model.


\subsection{User Behavior Sequence Modeling in Recommendation Systems}

Sequence modeling has been widely used in e-Commerce or in the prediction of CTR in ads~\cite{ZhouZSFZMYJLG18,FengLSWSZY19,abs-1809-03672,abs-1905-09248,abs-2006-05639,abs-2205-10249,ChangZFZGLHLNSG23}. Another category of modeling user interest is to enrich the input data~\cite{ZhengZ0C22, JiangJWLGZSXZ22,FanOGFLBDZZL22}. These papers often use a variant of Transformer encoder or the attention mechanism to implement the forward user or item encoding. However, they are all built for CTR prediction, instead of CVR prediction. Therefore, they do not consider the label sparseness. The techniques in existing works that optimize long or life-time sequence are not necessary for our purposes because the off-site sequence is usually short.

However, the trend in technique in these works on how to leverage the input data is inspirational to our future works, including: 1) use long user behavior sequence. TWIN~\cite{ChangZFZGLHLNSG23} proposed a general retrieval component to construct a shorter relevant sequence first. This unlocks the inference efficiency of using a life-long user behavior sequence; 2) enrich per item information in the sequence; 3) user more complex data structure than sequence.

\subsection{CL, SSL, and MTL}

The auxiliary objective function of our multisequence modeling is a type of CL~\cite{abs-2011-00362}, which tries to push a positive off-site event far away from randomly sampled negative events in the embedding space. Because the auxiliary loss is constructed from the input sequence feature, instead of the conversion label, it is essentially an SSL method~\cite{YuYXCLH24}. Finally, because the master conversion prediction model learns multiple conversion events simultaneously, and the SSL loss is also added to the supervised loss, our learning method is MTL~\cite{abs-2302-03525}.

This combination of CL, SSL and MTL is able to help the main learning task, conversion rate prediction, by exploring the user's interests contained in the sequence features. We are essentially formulating an autoregression problem as predicting the item in the attributed conversion sequence. It augments the labels by the past conversions per user. 

\section{Experiments and Results}\label{sec:exp}

We present the empirical results and learnings on the practice of CVR prediction, biased towards the models that have been working in real production.

\subsection{Dataset}

Our experiments are conducted on a sample of Pinterest's large-scale production dataset, which includes:

\begin{itemize}
    \item \textbf{Positives}: All off-site ad insertions attributed conversions (that is, checkout) with full engagement labels, such as clicks and repins.
    \item \textbf{Negatives}: A downsampled set of 5\% insertions without attributed conversion.
    
\end{itemize}

The sampled dataset has approximately hundreds of million samples and tens of millions of unique users per day. The training periods for our models range from 60 to 150 days, ensuring sufficient data for robust model training and evaluation. PyTorch DistributedDataParallel is used in the trainer. Training data is randomly shuffled. The training mechanism follows a batch + incremental mode, where we train a warm-up model with 110 days of data. Then we train incrementally for 40 days loading the previous day's checkpoint to warm up.

\subsection{Evaluation Setup and Metrics}




We evaluated the performance of the model on the next day and calculated the average performance. Because CVR prediction is essentially a classification problem, we use ROC-AUC~\cite{ROC-AUC} and Precision-Recall AUC (PR-AUC)~\cite{PR-AUC} as the metric. PR-AUC is supposed to be better because positive: negative $\ll 1$ is highly skewed. However, we found that either of them can be a better metric than the other when evaluated in an online ad recommendation system.

Average Cost Per Action (CPA) is the online evaluation metric, which is a standard metric in the advertising industry, calculated by dividing the total cost of conversions by the total number of conversions. For example, if one ad receives 2 conversions, one costing \$2.00 and one costing \$4.00, then the average CPA for those conversions is $(\$2.00 + \$4.00)/2=\$3.00$. The smaller the CPA on an ad platform, the more competitive and attractive its advertising effect for advertisers.

\subsection{From Single Feature-Crossing Module to Ensemble Module}

The first question we examine is whether an ensemble model is better than a single feature-crossing module. The baseline method is MLP as a single crossing module. This method is actually a strong baseline that has been in production for years. Starting from it, we tried many modules developed in recent years. Depending on the data set on which we train and evaluate, not all modules can lead to significant gains. Specifically, we found that the following four modules, among many architectures proposed for CTR prediction, are most effective for CVR prediction both offline and online:

\begin{itemize}
    \item MLP: Multi-layer Perceptron as the baseline feature crossing module;
    \item DCN V2: Improved DCN with low-rank optimization~\cite{WangSCJLHC21};
    \item MaskNet: Similar to DCN V2 but have parameters with better normalization~\cite{abs-2102-07619};
    \item Transformer: Standard version in ~\cite{VaswaniSPUJGKP17};
    \item DHEN: Two-layer ensemble model~\cite{abs-2203-11014}.
\end{itemize}

\begin{figure}
\vspace{-0.3cm}
\centering
\includegraphics[width=\columnwidth]{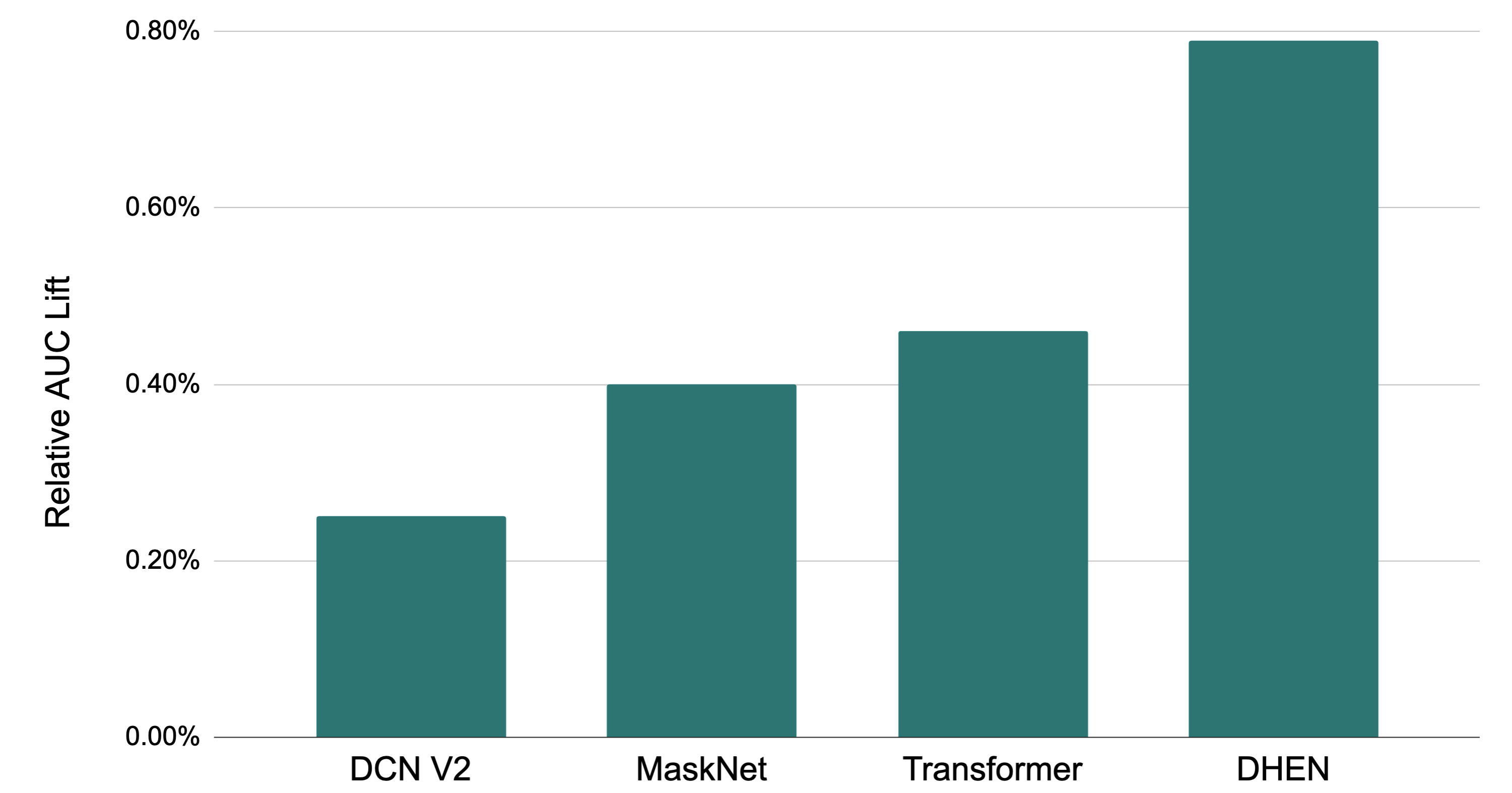}
\vspace{-0.4cm}
\caption{The best offline relative ROC-AUC lift of feature-crossing models compared to MLP baseline.}
\label{fig:lift}
\vspace{-0.5cm}
\end{figure}

\textbf{Offline Evaluation of Feature-Crossing Modules.} Figure~\ref{fig:lift} presents the relative lift in AUC compared to MLP. All 3 single feature-crossing modules show clear offline gains. DHEN is the best by a large margin, with the same set of input features. This verifies that the ensemble model is capable of capturing the effectiveness of each crossing module. It can bring additional benefits and has been one of the most successful launches in production.

\textbf{Ablation on DHEN configurations.} Due to the flexibility of DHEN, it is not clear how deep and how wide DHEN can lead to the best performance. In each layer, it can use any of the combinations of single feature-crossing modules. To evaluate this, using the best Transformer model as the baseline, we summarize the best DHEN tuning results with the following schema:

\begin{itemize}
    \item $\{1, 2, 3\}$-Layer: DHEN with different number of layers;
    \item 3-Layer-S: 3-layer DHEN use exactly the same crossing module per layer
\end{itemize}

\begin{table}
 \center
  \caption{Best DHEN of different number of layers and different feature crossing modules per layer. \checkmark means a module is used on that layer, and \ding{53} means a module is not used.}
  \vspace{-0.4cm}
  \begin{tabular}{rccccc}
    \toprule
    &MLP &DCN V2 &MaskNet &Trans &AUC Lift$^\S$ \\
    \toprule

    \multirow{1}{*}[0cm]{\begin{tabular}{r}1-Layer\end{tabular}} &\checkmark &\ding{53} &\ding{53} &\ding{53} &-0.18\%\\
    \midrule
    
    \multirow{2}{*}[0cm]{\begin{tabular}{r}2-Layer\end{tabular}}
    &\checkmark &\checkmark &\ding{53} &\checkmark   &\multirow{2}{*}[0cm]{\begin{tabular}{c}0.31\%\end{tabular}} \\
    \cmidrule{2-5}
    &\checkmark &\checkmark &\checkmark &\checkmark  \\
    \midrule
     
    \multirow{3}{*}[0cm]{\begin{tabular}{r}3-Layer\end{tabular}}
    &\checkmark &\checkmark &\ding{53} &\checkmark   &\multirow{3}{*}[-0.1cm]{\begin{tabular}{c}0.30\%\end{tabular}} \\
    \cmidrule{2-5}
    &\checkmark &\checkmark &\checkmark &\checkmark  \\
    \cmidrule{2-5}
    &\checkmark &\checkmark &\checkmark &\checkmark \\
    \midrule
    
    3-Layer-S &\checkmark &\checkmark &\ding{53} &\checkmark   &0.08\% \\
    
    \bottomrule
  \end{tabular}
\S: Baseline is 4 layer Transformer as a single crossing module.
\label{table:dhen}
\vspace{-0.4cm}
\end{table}

Table~\ref{table:dhen} presents the best AUC result for each setup using ParetoNet training. We draw several interesting observations.

First, the best tuned single layer DHEN cannot beat the best multi-layer Transformer. Instead, it drops the AUC by 0. 18\%. This means that going to multiple layers is still important for DHEN. It does not necessarily beat a carefully tuned single-crossing module.

Second, there is no essential difference after two layers for DHEN, while the inference cost will increase. We deployed 2-Layer DHEN into production. Note that the MaskNet module in DHEN already contains 2 horizontal blocks, and the Transformer module inside DHEN already contains 2 layers with 4 heads, which means that the complexity of DHEN increases very quickly with the number of layers. We also found that training more epoches hurts 3-Layer DHEN's performance further, which means that over-fitting is probably the reason that more layers will not improve generalization performance any more.

However, the gain from 3-Layer-S is marginal. Fixing the modules in each layer diminishes the gains of DHEN. That confirms that different crossing modules can help each other.

\textbf{Online CPA Reduction.} The offline AUC metric does not necessarily map to online performance, because it couples with the ad delivery system with a large number of moving knobs. In order to make online results reliable, we also look at other important business metrics in addition to CPA. The online experiment fixes the ad campaign budget between the control group and the treatment group. The hypothesis for treatment is that the revenue should be close to neutral, but the CPA will decrease due to a better ranking result based on the CVR prediction.

\begin{table}
 \vspace{-0.3cm}
 \center
  \caption{Business metric change in the treatment group during online A/B test for deploying DHEN.}
  \vspace{-0.4cm}
  \begin{tabular}{rrlc}
    \toprule
    &Metric &Metric &Stats Sig\\
    \toprule
    \multirow{6}{*}[0cm]{\begin{tabular}{r}Advertiser\end{tabular}} &\textbf{CPA} &\textbf{-2.15\%} &\checkmark\\

    &\textbf{\#Conversion} &\textbf{+1.62\%} &\checkmark\\
    
    &CPC &-0.12\% &\ding{53}\\
    
    &CPM &+0.06\% &\ding{53}\\
    
    &Outbound Click &-0.39\% &\ding{53}\\
    
    &Overall \#Click &-3.37\% &\checkmark\\
    
    &Conversion \#Click &+0.44\% &\checkmark\\
    
    \midrule

    \multirow{6}{*}[0cm]{\begin{tabular}{r}Platform\end{tabular}} &\textbf{iCVR} &\textbf{+4.90\%} &\checkmark\\
    &Revenue &-0.10\% &\ding{53}\\
    
    &Overall CTR &-2.70\% &\checkmark\\
    
    &Conv CTR &+0.08\% &\ding{53}\\
    
    &Ad Diversity &-0.87\% &\checkmark\\
    
    &Infra Cost &+\$1.2M/Year &\checkmark\\
    \midrule

    \multirow{4}{*}[0cm]{\begin{tabular}{r}User\end{tabular}} &Ad Impression &-0.68\% &\checkmark\\
    &Overall gCTR30 &-1.18\% &\checkmark\\
    &Conversion gCTR30 &-0.48\% &\ding{53}\\
    &Hide Rate &-0.13\% &\ding{53}\\
    
    \bottomrule
  \end{tabular}
\label{table:online}
\vspace{-0.6cm}
\end{table}

Table~\ref{table:online} presents the changes in the online metrics. We explain the meaning of each one and draw conclusions about its movement. For the most important business metric for oCPM ads, the CPA is significantly reduced by 2. 15\%, while the total number of conversions (\#conversion) increases by 1. 62\%, and the conversion rate per impression of ads (iCVR) increases by 4.90\%. This is a great success for a model update. We reduce the cost for advertisers to make platform's business more competitive. And it improves conversion volume at the same time.

From a platform perspective, the overall CTR drops by 2.70\%. However, note that this treatment group works only on the oCPM ad. The conversion ad CTR is actually neutral (0.08\% statistically not significant). This learning is important: For oCPM ad campaigns, a significant drop in CTR but a significant increase in conversion volume can occur at the same time. The implication is strong in driving future directions:
\begin{itemize}
    \item Business Wise: click on ad can be from entertainment purpose, it does not necessarily provide clear conversion value;
    \item Model Wise: entire space optimization can be promising, which means we optimize CTR and CVR together, instead of depending on engagement models predicting CTR.
\end{itemize}

From a user experience perspective, the overall ad load drops by 0. 68\%, which is usually a positive move. gCTR30 means "good click lasting 30 seconds", which follows similar trends to CTR. The change in ad hide rate is negligible.

In summary, our practice leads to a significantly better CVR prediction model and increases advertiser value and platform value by a clear margin, without hurting user experience.

\subsection{Ablation of the Self-Supervised Loss}

Using the ParetoNet search of the parameters in Section~\ref{sec:pareto}, we are able to identify a few configurations for self-supervised loss, to gain insight into the importance of different factors. Specifically, we examine three areas:
\begin{itemize}
    \item Prediction Goal: We predict only the future actions in the input sequences (Next Action Loss abbreviated as NAL), versus the randomly masked actions in the middle (Masked Language Modeling abbreviated as MLM);
    \item Data Sampling Ratio: how many positive actions (\#Pos) to predict and how many negative actions (\#Neg) to sample per positive;
    \item Importance of SSL: we can put different weight of the SSL term, split by organic sequences (OrgWeight) and ads sequences (AdsWeight).
\end{itemize}

\begin{table}
 \center
  \caption{Ablation study of factors impacting SSL performance.}
  \vspace{-0.4cm}
  \begin{tabular}{rccccc}
    \toprule
    &\#Pos	&\#Neg	&OrgWeight	&AdsWeight	&AUC Lift \\
    \toprule
    NAL &90	&20	&0.0002	&0.0001	&2.18\%\\
    NAL &90	&20	&0	&0.0002	&0.90\%\\
    \midrule
    NAL &20	&100	&0.0005	&0.0005	&1.03\%\\
    NAL &20	&100	&0.001	&0.001	&0.65\%\\
    NAL &20	&100	&0.01	&0.01	&0.29\%\\
    \midrule
    MLM &60	&30	&0.0002	&0.0001	&1.68\%\\    
    \bottomrule
  \end{tabular}
\label{table:nal}
\vspace{-0.5cm}
\end{table}

Table~\ref{table:nal} presents the best ParetoNet results for PR-AUC with different configurations. First, the best result is achieved by NAL prediction, which means that predicting future actions and the most recent actions help more than predicting past actions. In terms of SSL data sampling, it is clear that more NAL predictions are more useful (i.e., \#Pos is bigger). But different from pre-trained embedding learning, the ratio of negative per positive does not need to be a large number. A \#Pos: \#Neg = 1:20 produces the best results. This is probably explained by the fact that NAL is just a helper term in an objective function.


\subsection{Ablation of Different Category of User Features}

The user-side features are very important for the personalized ad recommendation. Although we derived a better model with DHEN, we hope to examine which types of features are most effective for the prediction. To this end, we divide the user-side features into 5 categories: \textit{Demographic} (e.g., gender, age, location), \textit{Counting} (e.g., onsite and off-site action count in a past time window), \textit{Categorical} (e.g., predicted user interest, annotation), \textit{Pre-trained User Embedding}, and \textit{Sequence}.

\begin{figure}
\centering
\includegraphics[width=\columnwidth]{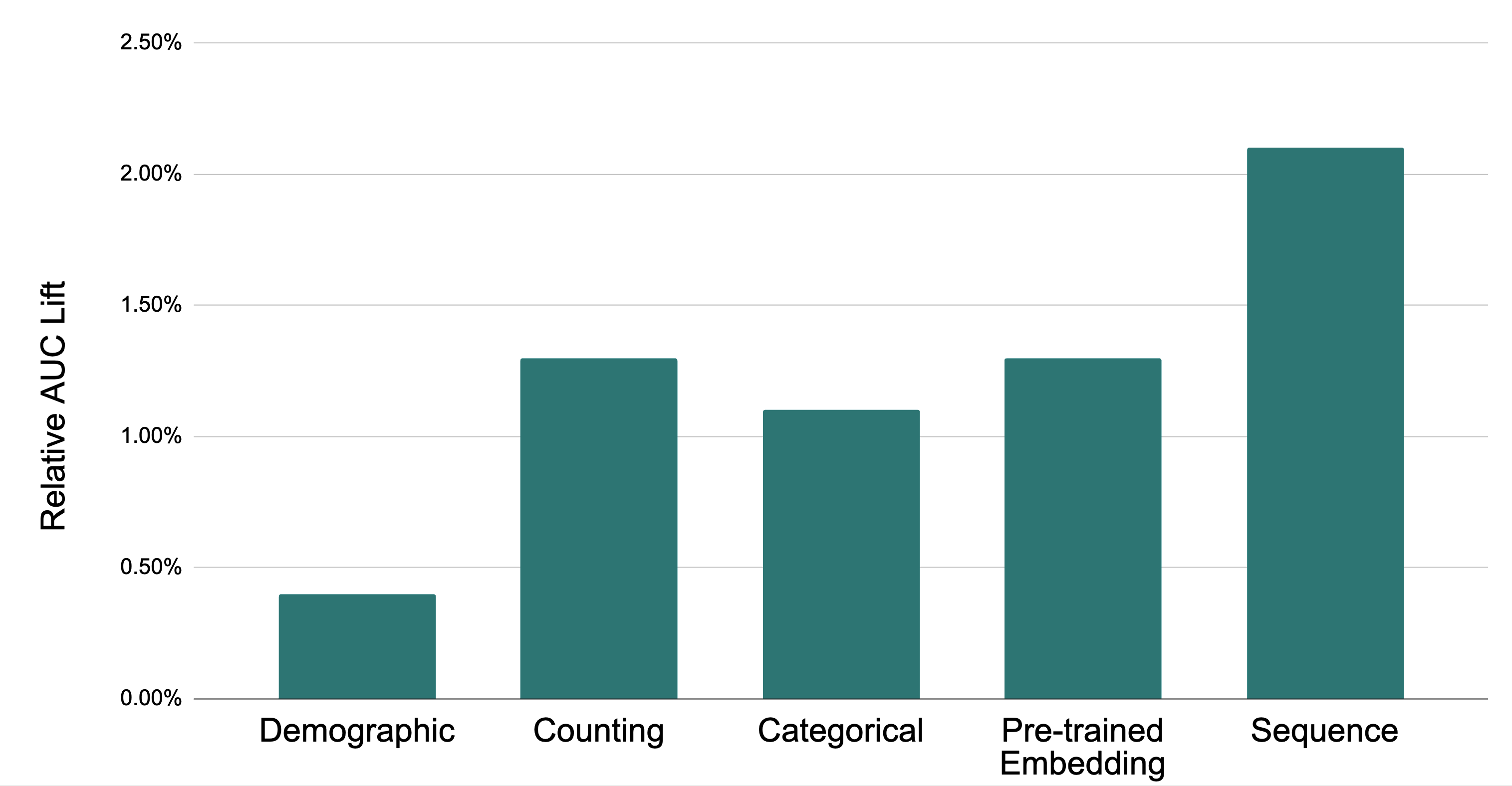}
\vspace{-0.7cm}
\caption{The relative AUC lift of adding each category of user features, with a baseline of removing all user-side features.}
\label{fig:feature-lift}
\vspace{-0.5cm}
\end{figure}

We completely remove all user-side features as a baseline. Then we evaluate AUC lift by adding a single category of user-side features. Figure~\ref{fig:feature-lift} presents the AUC gains for each type of feature. Sequence features are most effective, which double confirms the effect of SSL based user modeling. Another important observation is that pre-trained embedding cannot replace other features, although they are often included in the input of pre-trained embedding learning. The most possible reason we believe is that pre-trained embedding has a different objective function from the CVR prediction task.

\section{Conclusion and Future Works}\label{sec:exp}

The prediction of the conversion rate is of the utmost importance in an online advertising platform. We proposed applying an ensemble approach for the ad conversion rate prediction with an auxilary self-supervised future action prediction task for the input sequences, biased to the applied data science perspective on what factors are most effective for the final model's performance. We conducted a detailed ablation study on both hyperparameters and user-side features. The proposed modeling approach achieved great offline gains and online business metric gains. Looking into the future, there are four important areas that call for cutting-edge applied science and techniques.

1) Entire space optimization: the CVR prediction is orthogonal to the engagement models in reality. It is shown that CVR can increase while CTR drops. Optimize engagement tasks such as CTR, gCTR30, favorite, etc. can probably improve the performance on the CVR. 2) Better sequence modeling: The user behavior sequence turns out to be the most important user-side feature. It deserves more exploration on how to maximize its effect, including even longer sequences, a generative AI modeling approach, and enriching the items in sequence. 3) Delayed feedback of conversion: CVR labels depend on the measurement of conversion in the third-party data. It usually has a delay in collecting this data. 4) Privacy-preserving machine learning: a trend in industry is to protect user data which will have a clear negative impact on the quantity and quality of training data. How to optimize models while respecting users' privacy is an important topic for oCPM ads.

\bibliographystyle{ACM-Reference-Format}
\bibliography{dhen_bib}


\begin{thebibliography}{41}


\ifx \showCODEN    \undefined \def \showCODEN     #1{\unskip}     \fi
\ifx \showDOI      \undefined \def \showDOI       #1{#1}\fi
\ifx \showISBNx    \undefined \def \showISBNx     #1{\unskip}     \fi
\ifx \showISBNxiii \undefined \def \showISBNxiii  #1{\unskip}     \fi
\ifx \showISSN     \undefined \def \showISSN      #1{\unskip}     \fi
\ifx \showLCCN     \undefined \def \showLCCN      #1{\unskip}     \fi
\ifx \shownote     \undefined \def \shownote      #1{#1}          \fi
\ifx \showarticletitle \undefined \def \showarticletitle #1{#1}   \fi
\ifx \showURL      \undefined \def \showURL       {\relax}        \fi
\providecommand\bibfield[2]{#2}
\providecommand\bibinfo[2]{#2}
\providecommand\natexlab[1]{#1}
\providecommand\showeprint[2][]{arXiv:#2}

\bibitem[Cao et~al\mbox{.}(2022)]%
        {abs-2205-10249}
\bibfield{author}{\bibinfo{person}{Yue Cao}, \bibinfo{person}{Xiaojiang Zhou}, \bibinfo{person}{Jiaqi Feng}, \bibinfo{person}{Peihao Huang}, \bibinfo{person}{Yao Xiao}, \bibinfo{person}{Dayao Chen}, {and} \bibinfo{person}{Sheng Chen}.} \bibinfo{year}{2022}\natexlab{}.
\newblock \showarticletitle{Sampling Is All You Need on Modeling Long-Term User Behaviors for {CTR} Prediction}.
\newblock \bibinfo{journal}{\emph{CoRR}}  \bibinfo{volume}{abs/2205.10249} (\bibinfo{year}{2022}).
\newblock
\showeprint[arXiv]{2205.10249}
\urldef\tempurl%
\url{https://doi.org/10.48550/arXiv.2205.10249}
\showURL{%
\tempurl}


\bibitem[Chang et~al\mbox{.}(2023)]%
        {ChangZFZGLHLNSG23}
\bibfield{author}{\bibinfo{person}{Jianxin Chang}, \bibinfo{person}{Chenbin Zhang}, \bibinfo{person}{Zhiyi Fu}, \bibinfo{person}{Xiaoxue Zang}, \bibinfo{person}{Lin Guan}, \bibinfo{person}{Jing Lu}, \bibinfo{person}{Yiqun Hui}, \bibinfo{person}{Dewei Leng}, \bibinfo{person}{Yanan Niu}, \bibinfo{person}{Yang Song}, {and} \bibinfo{person}{Kun Gai}.} \bibinfo{year}{2023}\natexlab{}.
\newblock \showarticletitle{{TWIN:} TWo-stage Interest Network for Lifelong User Behavior Modeling in {CTR} Prediction at Kuaishou}. In \bibinfo{booktitle}{\emph{DKK}}. \bibinfo{pages}{3785--3794}.
\newblock


\bibitem[Davis and Goadrich(2006)]%
        {PR-AUC}
\bibfield{author}{\bibinfo{person}{Jesse Davis} {and} \bibinfo{person}{Mark Goadrich}.} \bibinfo{year}{2006}\natexlab{}.
\newblock \showarticletitle{The relationship between Precision-Recall and {ROC} curves}. In \bibinfo{booktitle}{\emph{ICML}}, Vol.~\bibinfo{volume}{148}. \bibinfo{pages}{233--240}.
\newblock


\bibitem[deWet and Ou(2019)]%
        {deWetO19}
\bibfield{author}{\bibinfo{person}{Stephanie deWet} {and} \bibinfo{person}{Jiafan Ou}.} \bibinfo{year}{2019}\natexlab{}.
\newblock \showarticletitle{Finding Users Who Act Alike: Transfer Learning for Expanding Advertiser Audiences}. In \bibinfo{booktitle}{\emph{KDD}}. \bibinfo{pages}{2251--2259}.
\newblock


\bibitem[Fan et~al\mbox{.}(2022)]%
        {FanOGFLBDZZL22}
\bibfield{author}{\bibinfo{person}{Zhifang Fan}, \bibinfo{person}{Dan Ou}, \bibinfo{person}{Yulong Gu}, \bibinfo{person}{Bairan Fu}, \bibinfo{person}{Xiang Li}, \bibinfo{person}{Wentian Bao}, \bibinfo{person}{Xin{-}Yu Dai}, \bibinfo{person}{Xiaoyi Zeng}, \bibinfo{person}{Tao Zhuang}, {and} \bibinfo{person}{Qingwen Liu}.} \bibinfo{year}{2022}\natexlab{}.
\newblock \showarticletitle{Modeling Users' Contextualized Page-wise Feedback for Click-Through Rate Prediction in E-commerce Search}. In \bibinfo{booktitle}{\emph{WSDM}}. \bibinfo{pages}{262--270}.
\newblock


\bibitem[Feng et~al\mbox{.}(2019)]%
        {FengLSWSZY19}
\bibfield{author}{\bibinfo{person}{Yufei Feng}, \bibinfo{person}{Fuyu Lv}, \bibinfo{person}{Weichen Shen}, \bibinfo{person}{Menghan Wang}, \bibinfo{person}{Fei Sun}, \bibinfo{person}{Yu Zhu}, {and} \bibinfo{person}{Keping Yang}.} \bibinfo{year}{2019}\natexlab{}.
\newblock \showarticletitle{Deep Session Interest Network for Click-Through Rate Prediction}. In \bibinfo{booktitle}{\emph{IJCAI}}. \bibinfo{pages}{2301--2307}.
\newblock


\bibitem[Grbovic and Cheng(2018)]%
        {GrbovicC18}
\bibfield{author}{\bibinfo{person}{Mihajlo Grbovic} {and} \bibinfo{person}{Haibin Cheng}.} \bibinfo{year}{2018}\natexlab{}.
\newblock \showarticletitle{Real-time Personalization using Embeddings for Search Ranking at Airbnb}. In \bibinfo{booktitle}{\emph{KDD}}, \bibfield{editor}{\bibinfo{person}{Yike Guo} {and} \bibinfo{person}{Faisal Farooq}} (Eds.). \bibinfo{publisher}{{ACM}}, \bibinfo{pages}{311--320}.
\newblock


\bibitem[Guo et~al\mbox{.}(2017)]%
        {GuoTYLH17}
\bibfield{author}{\bibinfo{person}{Huifeng Guo}, \bibinfo{person}{Ruiming Tang}, \bibinfo{person}{Yunming Ye}, \bibinfo{person}{Zhenguo Li}, {and} \bibinfo{person}{Xiuqiang He}.} \bibinfo{year}{2017}\natexlab{}.
\newblock \showarticletitle{DeepFM: {A} Factorization-Machine based Neural Network for {CTR} Prediction}. In \bibinfo{booktitle}{\emph{IJCAI}}, \bibfield{editor}{\bibinfo{person}{Carles Sierra}} (Ed.). \bibinfo{pages}{1725--1731}.
\newblock


\bibitem[Hamilton et~al\mbox{.}(2017)]%
        {HamiltonYL17}
\bibfield{author}{\bibinfo{person}{William~L. Hamilton}, \bibinfo{person}{Zhitao Ying}, {and} \bibinfo{person}{Jure Leskovec}.} \bibinfo{year}{2017}\natexlab{}.
\newblock \showarticletitle{Inductive Representation Learning on Large Graphs}. In \bibinfo{booktitle}{\emph{NeurIPS}}. \bibinfo{pages}{1024--1034}.
\newblock


\bibitem[Hanley and McNeil(1982)]%
        {ROC-AUC}
\bibfield{author}{\bibinfo{person}{James~A. Hanley} {and} \bibinfo{person}{Barbara~J. McNeil}.} \bibinfo{year}{1982}\natexlab{}.
\newblock \showarticletitle{The meaning and use of the area under a receiver operating characteristic (ROC) curve}.
\newblock \bibinfo{journal}{\emph{Radiology}}  \bibinfo{volume}{143(1)} (\bibinfo{year}{1982}), \bibinfo{pages}{29--36}.
\newblock


\bibitem[Hsu et~al\mbox{.}(2024)]%
        {hsu2024taming}
\bibfield{author}{\bibinfo{person}{Yi-Ping Hsu}, \bibinfo{person}{Po-Wei Wang}, \bibinfo{person}{Chantat Eksombatchai}, {and} \bibinfo{person}{Jiajing Xu}.} \bibinfo{year}{2024}\natexlab{}.
\newblock \showarticletitle{Taming the One-Epoch Phenomenon in Online Recommendation System by Two-stage Contrastive ID Pre-training}. In \bibinfo{booktitle}{\emph{Proceedings of the 18th ACM Conference on Recommender Systems}}. \bibinfo{pages}{838--840}.
\newblock


\bibitem[Hu et~al\mbox{.}(2018)]%
        {HuSS18}
\bibfield{author}{\bibinfo{person}{Jie Hu}, \bibinfo{person}{Li Shen}, {and} \bibinfo{person}{Gang Sun}.} \bibinfo{year}{2018}\natexlab{}.
\newblock \showarticletitle{Squeeze-and-Excitation Networks}. In \bibinfo{booktitle}{\emph{CVPR}}. \bibinfo{pages}{7132--7141}.
\newblock


\bibitem[Huang et~al\mbox{.}(2019)]%
        {huang2019fibinet}
\bibfield{author}{\bibinfo{person}{Tongwen Huang}, \bibinfo{person}{Zhiqi Zhang}, {and} \bibinfo{person}{Junlin Zhang}.} \bibinfo{year}{2019}\natexlab{}.
\newblock \showarticletitle{FiBiNET: combining feature importance and bilinear feature interaction for click-through rate prediction}. In \bibinfo{booktitle}{\emph{Proceedings of the 13th ACM conference on recommender systems}}. \bibinfo{pages}{169--177}.
\newblock


\bibitem[Jaiswal et~al\mbox{.}(2020)]%
        {abs-2011-00362}
\bibfield{author}{\bibinfo{person}{Ashish Jaiswal}, \bibinfo{person}{Ashwin~Ramesh Babu}, \bibinfo{person}{Mohammad~Zaki Zadeh}, \bibinfo{person}{Debapriya Banerjee}, {and} \bibinfo{person}{Fillia Makedon}.} \bibinfo{year}{2020}\natexlab{}.
\newblock \showarticletitle{A Survey on Contrastive Self-supervised Learning}.
\newblock \bibinfo{journal}{\emph{CoRR}}  \bibinfo{volume}{abs/2011.00362} (\bibinfo{year}{2020}).
\newblock
\showeprint[arXiv]{2011.00362}


\bibitem[Jiang et~al\mbox{.}(2022)]%
        {JiangJWLGZSXZ22}
\bibfield{author}{\bibinfo{person}{Wensen Jiang}, \bibinfo{person}{Yizhu Jiao}, \bibinfo{person}{Qingqin Wang}, \bibinfo{person}{Chuanming Liang}, \bibinfo{person}{Lijie Guo}, \bibinfo{person}{Yao Zhang}, \bibinfo{person}{Zhijun Sun}, \bibinfo{person}{Yun Xiong}, {and} \bibinfo{person}{Yangyong Zhu}.} \bibinfo{year}{2022}\natexlab{}.
\newblock \showarticletitle{Triangle Graph Interest Network for Click-through Rate Prediction}. In \bibinfo{booktitle}{\emph{WSDM}}. \bibinfo{pages}{401--409}.
\newblock


\bibitem[Li et~al\mbox{.}(2024)]%
        {li2024privacy}
\bibfield{author}{\bibinfo{person}{Kungang Li}, \bibinfo{person}{Xiangyi Chen}, \bibinfo{person}{Ling Leng}, \bibinfo{person}{Jiajing Xu}, \bibinfo{person}{Jiankai Sun}, {and} \bibinfo{person}{Behnam Rezaei}.} \bibinfo{year}{2024}\natexlab{}.
\newblock \showarticletitle{Privacy Preserving Conversion Modeling in Data Clean Room}. In \bibinfo{booktitle}{\emph{Proceedings of the 18th ACM Conference on Recommender Systems}}. \bibinfo{pages}{819--822}.
\newblock


\bibitem[Pal et~al\mbox{.}(2020)]%
        {PalEZZRL20}
\bibfield{author}{\bibinfo{person}{Aditya Pal}, \bibinfo{person}{Chantat Eksombatchai}, \bibinfo{person}{Yitong Zhou}, \bibinfo{person}{Bo Zhao}, \bibinfo{person}{Charles Rosenberg}, {and} \bibinfo{person}{Jure Leskovec}.} \bibinfo{year}{2020}\natexlab{}.
\newblock \showarticletitle{PinnerSage: Multi-Modal User Embedding Framework for Recommendations at Pinterest}. In \bibinfo{booktitle}{\emph{KDD}}. \bibinfo{pages}{2311--2320}.
\newblock


\bibitem[Pancha et~al\mbox{.}(2022)]%
        {PanchaZLR22}
\bibfield{author}{\bibinfo{person}{Nikil Pancha}, \bibinfo{person}{Andrew Zhai}, \bibinfo{person}{Jure Leskovec}, {and} \bibinfo{person}{Charles Rosenberg}.} \bibinfo{year}{2022}\natexlab{}.
\newblock \showarticletitle{PinnerFormer: Sequence Modeling for User Representation at Pinterest}. In \bibinfo{booktitle}{\emph{KDD}}. \bibinfo{publisher}{{ACM}}, \bibinfo{pages}{3702--3712}.
\newblock


\bibitem[Pi et~al\mbox{.}(2019)]%
        {abs-1905-09248}
\bibfield{author}{\bibinfo{person}{Qi Pi}, \bibinfo{person}{Weijie Bian}, \bibinfo{person}{Guorui Zhou}, \bibinfo{person}{Xiaoqiang Zhu}, {and} \bibinfo{person}{Kun Gai}.} \bibinfo{year}{2019}\natexlab{}.
\newblock \showarticletitle{Practice on Long Sequential User Behavior Modeling for Click-Through Rate Prediction}.
\newblock \bibinfo{journal}{\emph{CoRR}}  \bibinfo{volume}{abs/1905.09248} (\bibinfo{year}{2019}).
\newblock
\showeprint[arXiv]{1905.09248}


\bibitem[Pi et~al\mbox{.}(2020)]%
        {abs-2006-05639}
\bibfield{author}{\bibinfo{person}{Qi Pi}, \bibinfo{person}{Xiaoqiang Zhu}, \bibinfo{person}{Guorui Zhou}, \bibinfo{person}{Yujing Zhang}, \bibinfo{person}{Zhe Wang}, \bibinfo{person}{Lejian Ren}, \bibinfo{person}{Ying Fan}, {and} \bibinfo{person}{Kun Gai}.} \bibinfo{year}{2020}\natexlab{}.
\newblock \showarticletitle{Search-based User Interest Modeling with Lifelong Sequential Behavior Data for Click-Through Rate Prediction}.
\newblock \bibinfo{journal}{\emph{CoRR}}  \bibinfo{volume}{abs/2006.05639} (\bibinfo{year}{2020}).
\newblock
\showeprint[arXiv]{2006.05639}


\bibitem[Rosenblatt(1958)]%
        {rosenblatt1958perceptron}
\bibfield{author}{\bibinfo{person}{F. Rosenblatt}.} \bibinfo{year}{1958}\natexlab{}.
\newblock \showarticletitle{{The perceptron: A probabilistic model for information storage and organization in the brain.}}
\newblock \bibinfo{journal}{\emph{Psychological Review}} \bibinfo{volume}{65}, \bibinfo{number}{6} (\bibinfo{year}{1958}), \bibinfo{pages}{386--408}.
\newblock
\showISSN{0033-295X}
\urldef\tempurl%
\url{https://doi.org/10.1037/h0042519}
\showDOI{\tempurl}


\bibitem[van~den Oord et~al\mbox{.}(2018)]%
        {abs-1807-03748}
\bibfield{author}{\bibinfo{person}{A{\"{a}}ron van~den Oord}, \bibinfo{person}{Yazhe Li}, {and} \bibinfo{person}{Oriol Vinyals}.} \bibinfo{year}{2018}\natexlab{}.
\newblock \showarticletitle{Representation Learning with Contrastive Predictive Coding}.
\newblock \bibinfo{journal}{\emph{CoRR}}  \bibinfo{volume}{abs/1807.03748} (\bibinfo{year}{2018}).
\newblock
\showeprint[arXiv]{1807.03748}


\bibitem[Vaswani et~al\mbox{.}(2017)]%
        {VaswaniSPUJGKP17}
\bibfield{author}{\bibinfo{person}{Ashish Vaswani}, \bibinfo{person}{Noam Shazeer}, \bibinfo{person}{Niki Parmar}, \bibinfo{person}{Jakob Uszkoreit}, \bibinfo{person}{Llion Jones}, \bibinfo{person}{Aidan~N. Gomez}, \bibinfo{person}{Lukasz Kaiser}, {and} \bibinfo{person}{Illia Polosukhin}.} \bibinfo{year}{2017}\natexlab{}.
\newblock \showarticletitle{Attention is All you Need}. In \bibinfo{booktitle}{\emph{NeurIPS}}. \bibinfo{pages}{6000--6010}.
\newblock


\bibitem[Wang et~al\mbox{.}(2023a)]%
        {wang2023towards}
\bibfield{author}{\bibinfo{person}{Fangye Wang}, \bibinfo{person}{Hansu Gu}, \bibinfo{person}{Dongsheng Li}, \bibinfo{person}{Tun Lu}, \bibinfo{person}{Peng Zhang}, {and} \bibinfo{person}{Ning Gu}.} \bibinfo{year}{2023}\natexlab{a}.
\newblock \showarticletitle{Towards deeper, lighter and interpretable cross network for ctr prediction}. In \bibinfo{booktitle}{\emph{Proceedings of the 32nd ACM International Conference on Information and Knowledge Management}}. \bibinfo{pages}{2523--2533}.
\newblock


\bibitem[Wang et~al\mbox{.}(2017)]%
        {WangFFW17}
\bibfield{author}{\bibinfo{person}{Ruoxi Wang}, \bibinfo{person}{Bin Fu}, \bibinfo{person}{Gang Fu}, {and} \bibinfo{person}{Mingliang Wang}.} \bibinfo{year}{2017}\natexlab{}.
\newblock \showarticletitle{Deep {\&} Cross Network for Ad Click Predictions}. In \bibinfo{booktitle}{\emph{ADKDD}}. \bibinfo{pages}{12:1--12:7}.
\newblock


\bibitem[Wang et~al\mbox{.}(2021b)]%
        {WangSCJLHC21}
\bibfield{author}{\bibinfo{person}{Ruoxi Wang}, \bibinfo{person}{Rakesh Shivanna}, \bibinfo{person}{Derek~Zhiyuan Cheng}, \bibinfo{person}{Sagar Jain}, \bibinfo{person}{Dong Lin}, \bibinfo{person}{Lichan Hong}, {and} \bibinfo{person}{Ed~H. Chi}.} \bibinfo{year}{2021}\natexlab{b}.
\newblock \showarticletitle{{DCN} {V2:} Improved Deep {\&} Cross Network and Practical Lessons for Web-scale Learning to Rank Systems}. In \bibinfo{booktitle}{\emph{WWW}}. \bibinfo{pages}{1785--1797}.
\newblock


\bibitem[Wang et~al\mbox{.}(2023b)]%
        {abs-2302-03525}
\bibfield{author}{\bibinfo{person}{Yuhao Wang}, \bibinfo{person}{Ha~Tsz Lam}, \bibinfo{person}{Yi Wong}, \bibinfo{person}{Ziru Liu}, \bibinfo{person}{Xiangyu Zhao}, \bibinfo{person}{Yichao Wang}, \bibinfo{person}{Bo Chen}, \bibinfo{person}{Huifeng Guo}, {and} \bibinfo{person}{Ruiming Tang}.} \bibinfo{year}{2023}\natexlab{b}.
\newblock \showarticletitle{Multi-Task Deep Recommender Systems: {A} Survey}.
\newblock \bibinfo{journal}{\emph{CoRR}}  \bibinfo{volume}{abs/2302.03525} (\bibinfo{year}{2023}).
\newblock
\showeprint[arXiv]{2302.03525}


\bibitem[Wang et~al\mbox{.}(2021a)]%
        {abs-2102-07619}
\bibfield{author}{\bibinfo{person}{Zhiqiang Wang}, \bibinfo{person}{Qingyun She}, {and} \bibinfo{person}{Junlin Zhang}.} \bibinfo{year}{2021}\natexlab{a}.
\newblock \showarticletitle{MaskNet: Introducing Feature-Wise Multiplication to {CTR} Ranking Models by Instance-Guided Mask}.
\newblock \bibinfo{journal}{\emph{CoRR}}  \bibinfo{volume}{abs/2102.07619} (\bibinfo{year}{2021}).
\newblock
\showeprint[arXiv]{2102.07619}


\bibitem[Xia et~al\mbox{.}(2023)]%
        {XiaEPBWGJFZZ23}
\bibfield{author}{\bibinfo{person}{Xue Xia}, \bibinfo{person}{Pong Eksombatchai}, \bibinfo{person}{Nikil Pancha}, \bibinfo{person}{Dhruvil~Deven Badani}, \bibinfo{person}{Po{-}Wei Wang}, \bibinfo{person}{Neng Gu}, \bibinfo{person}{Saurabh~Vishwas Joshi}, \bibinfo{person}{Nazanin Farahpour}, \bibinfo{person}{Zhiyuan Zhang}, {and} \bibinfo{person}{Andrew Zhai}.} \bibinfo{year}{2023}\natexlab{}.
\newblock \showarticletitle{TransAct: Transformer-based Realtime User Action Model for Recommendation at Pinterest}. In \bibinfo{booktitle}{\emph{KDD}}. \bibinfo{publisher}{{ACM}}, \bibinfo{pages}{5249--5259}.
\newblock


\bibitem[Xu et~al\mbox{.}(2022)]%
        {XuZR22}
\bibfield{author}{\bibinfo{person}{Jiajing Xu}, \bibinfo{person}{Andrew Zhai}, {and} \bibinfo{person}{Charles Rosenberg}.} \bibinfo{year}{2022}\natexlab{}.
\newblock \showarticletitle{Rethinking Personalized Ranking at Pinterest: An End-to-End Approach}. In \bibinfo{booktitle}{\emph{RecSys '22}}. \bibinfo{publisher}{{ACM}}, \bibinfo{pages}{502--505}.
\newblock


\bibitem[Ying et~al\mbox{.}(2018)]%
        {YingHCEHL18}
\bibfield{author}{\bibinfo{person}{Rex Ying}, \bibinfo{person}{Ruining He}, \bibinfo{person}{Kaifeng Chen}, \bibinfo{person}{Pong Eksombatchai}, \bibinfo{person}{William~L. Hamilton}, {and} \bibinfo{person}{Jure Leskovec}.} \bibinfo{year}{2018}\natexlab{}.
\newblock \showarticletitle{Graph Convolutional Neural Networks for Web-Scale Recommender Systems}. In \bibinfo{booktitle}{\emph{KDD}}. \bibinfo{pages}{974--983}.
\newblock
\urldef\tempurl%
\url{https://doi.org/10.1145/3219819.3219890}
\showDOI{\tempurl}


\bibitem[Yu et~al\mbox{.}(2024)]%
        {YuYXCLH24}
\bibfield{author}{\bibinfo{person}{Junliang Yu}, \bibinfo{person}{Hongzhi Yin}, \bibinfo{person}{Xin Xia}, \bibinfo{person}{Tong Chen}, \bibinfo{person}{Jundong Li}, {and} \bibinfo{person}{Zi Huang}.} \bibinfo{year}{2024}\natexlab{}.
\newblock \showarticletitle{Self-Supervised Learning for Recommender Systems: {A} Survey}.
\newblock \bibinfo{journal}{\emph{{IEEE} Trans. Knowl. Data Eng.}} \bibinfo{volume}{36}, \bibinfo{number}{1} (\bibinfo{year}{2024}), \bibinfo{pages}{335--355}.
\newblock


\bibitem[Zhai et~al\mbox{.}(2019)]%
        {ZhaiWTPR19}
\bibfield{author}{\bibinfo{person}{Andrew Zhai}, \bibinfo{person}{Hao{-}Yu Wu}, \bibinfo{person}{Eric Tzeng}, \bibinfo{person}{Dong~Huk Park}, {and} \bibinfo{person}{Charles Rosenberg}.} \bibinfo{year}{2019}\natexlab{}.
\newblock \showarticletitle{Learning a Unified Embedding for Visual Search at Pinterest}. In \bibinfo{booktitle}{\emph{KDD}}. \bibinfo{pages}{2412--2420}.
\newblock


\bibitem[Zhai et~al\mbox{.}(2024)]%
        {abs-2402-17152}
\bibfield{author}{\bibinfo{person}{Jiaqi Zhai}, \bibinfo{person}{Lucy Liao}, \bibinfo{person}{Xing Liu}, \bibinfo{person}{Yueming Wang}, \bibinfo{person}{Rui Li}, \bibinfo{person}{Xuan Cao}, \bibinfo{person}{Leon Gao}, \bibinfo{person}{Zhaojie Gong}, \bibinfo{person}{Fangda Gu}, \bibinfo{person}{Michael He}, \bibinfo{person}{Yinghai Lu}, {and} \bibinfo{person}{Yu Shi}.} \bibinfo{year}{2024}\natexlab{}.
\newblock \showarticletitle{Actions Speak Louder than Words: Trillion-Parameter Sequential Transducers for Generative Recommendations}.
\newblock \bibinfo{journal}{\emph{CoRR}}  \bibinfo{volume}{abs/2402.17152} (\bibinfo{year}{2024}).
\newblock
\urldef\tempurl%
\url{https://doi.org/10.48550/ARXIV.2402.17152}
\showDOI{\tempurl}
\showeprint[arXiv]{2402.17152}


\bibitem[Zhang et~al\mbox{.}(2024)]%
        {zhang2024wukong}
\bibfield{author}{\bibinfo{person}{Buyun Zhang}, \bibinfo{person}{Liang Luo}, \bibinfo{person}{Yuxin Chen}, \bibinfo{person}{Jade Nie}, \bibinfo{person}{Xi Liu}, \bibinfo{person}{Daifeng Guo}, \bibinfo{person}{Yanli Zhao}, \bibinfo{person}{Shen Li}, \bibinfo{person}{Yuchen Hao}, \bibinfo{person}{Yantao Yao}, {et~al\mbox{.}}} \bibinfo{year}{2024}\natexlab{}.
\newblock \showarticletitle{Wukong: Towards a Scaling Law for Large-Scale Recommendation}.
\newblock \bibinfo{journal}{\emph{arXiv preprint arXiv:2403.02545}} (\bibinfo{year}{2024}).
\newblock


\bibitem[Zhang et~al\mbox{.}(2022)]%
        {abs-2203-11014}
\bibfield{author}{\bibinfo{person}{Buyun Zhang}, \bibinfo{person}{Liang Luo}, \bibinfo{person}{Xi Liu}, \bibinfo{person}{Jay Li}, \bibinfo{person}{Zeliang Chen}, \bibinfo{person}{Weilin Zhang}, \bibinfo{person}{Xiaohan Wei}, \bibinfo{person}{Yuchen Hao}, \bibinfo{person}{Michael Tsang}, \bibinfo{person}{Wenjun Wang}, \bibinfo{person}{Yang Liu}, \bibinfo{person}{Huayu Li}, \bibinfo{person}{Yasmine Badr}, \bibinfo{person}{Jongsoo Park}, \bibinfo{person}{Jiyan Yang}, \bibinfo{person}{Dheevatsa Mudigere}, {and} \bibinfo{person}{Ellie Wen}.} \bibinfo{year}{2022}\natexlab{}.
\newblock \showarticletitle{{DHEN:} {A} Deep and Hierarchical Ensemble Network for Large-Scale Click-Through Rate Prediction}.
\newblock \bibinfo{journal}{\emph{CoRR}}  \bibinfo{volume}{abs/2203.11014} (\bibinfo{year}{2022}).
\newblock


\bibitem[Zhang et~al\mbox{.}(2023)]%
        {zhang2023fibinet++}
\bibfield{author}{\bibinfo{person}{Pengtao Zhang}, \bibinfo{person}{Zheng Zheng}, {and} \bibinfo{person}{Junlin Zhang}.} \bibinfo{year}{2023}\natexlab{}.
\newblock \showarticletitle{FiBiNet++: Reducing model size by low rank feature interaction layer for CTR prediction}. In \bibinfo{booktitle}{\emph{Proceedings of the 32nd ACM International Conference on Information and Knowledge Management}}. \bibinfo{pages}{4425--4429}.
\newblock


\bibitem[Zheng et~al\mbox{.}(2022)]%
        {ZhengZ0C22}
\bibfield{author}{\bibinfo{person}{Zuowu Zheng}, \bibinfo{person}{Changwang Zhang}, \bibinfo{person}{Xiaofeng Gao}, {and} \bibinfo{person}{Guihai Chen}.} \bibinfo{year}{2022}\natexlab{}.
\newblock \showarticletitle{{HIEN:} Hierarchical Intention Embedding Network for Click-Through Rate Prediction}. In \bibinfo{booktitle}{\emph{SIGIR}}. \bibinfo{publisher}{{ACM}}, \bibinfo{pages}{322--331}.
\newblock


\bibitem[Zhou et~al\mbox{.}(2018a)]%
        {abs-1809-03672}
\bibfield{author}{\bibinfo{person}{Guorui Zhou}, \bibinfo{person}{Na Mou}, \bibinfo{person}{Ying Fan}, \bibinfo{person}{Qi Pi}, \bibinfo{person}{Weijie Bian}, \bibinfo{person}{Chang Zhou}, \bibinfo{person}{Xiaoqiang Zhu}, {and} \bibinfo{person}{Kun Gai}.} \bibinfo{year}{2018}\natexlab{a}.
\newblock \showarticletitle{Deep Interest Evolution Network for Click-Through Rate Prediction}.
\newblock \bibinfo{journal}{\emph{CoRR}}  \bibinfo{volume}{abs/1809.03672} (\bibinfo{year}{2018}).
\newblock
\showeprint[arXiv]{1809.03672}
\urldef\tempurl%
\url{http://arxiv.org/abs/1809.03672}
\showURL{%
\tempurl}


\bibitem[Zhou et~al\mbox{.}(2018b)]%
        {ZhouZSFZMYJLG18}
\bibfield{author}{\bibinfo{person}{Guorui Zhou}, \bibinfo{person}{Xiaoqiang Zhu}, \bibinfo{person}{Chengru Song}, \bibinfo{person}{Ying Fan}, \bibinfo{person}{Han Zhu}, \bibinfo{person}{Xiao Ma}, \bibinfo{person}{Yanghui Yan}, \bibinfo{person}{Junqi Jin}, \bibinfo{person}{Han Li}, {and} \bibinfo{person}{Kun Gai}.} \bibinfo{year}{2018}\natexlab{b}.
\newblock \showarticletitle{Deep Interest Network for Click-Through Rate Prediction}. In \bibinfo{booktitle}{\emph{KDD}}. \bibinfo{pages}{1059--1068}.
\newblock


\bibitem[Zhuang and Liu(2019)]%
        {ZhuangL19}
\bibfield{author}{\bibinfo{person}{Jinfeng Zhuang} {and} \bibinfo{person}{Yu Liu}.} \bibinfo{year}{2019}\natexlab{}.
\newblock \showarticletitle{PinText: {A} Multitask Text Embedding System in Pinterest}. In \bibinfo{booktitle}{\emph{KDD}}. \bibinfo{pages}{2653--2661}.
\newblock
\urldef\tempurl%
\url{https://doi.org/10.1145/3292500.3330671}
\showDOI{\tempurl}


\end{thebibliography}

\appendix

\section{Hyperparameters of the Deployed DHEN Model}

A significant amount of time was spent trying to determine which feature-crossing module to use in DHEN and what the parameters are for each module. In case it helps the audience, we list the parameters of each module in Table~\ref{table:hyperparameter}. We list the important parameter choices used in production below:
\begin{itemize}
    \item DHEN has two layers and each layer has 2 feature-crossing modules, where the ensemble modes are both sum;
    \item The final MLP module has 3 sequential layers $[128, 128, 128]$;
    \item The unified dimension projection module maps each feature to an embedding with dimension 64;
    \item User behavior sequence length is cut off at 500;
    \item The total number of model parameters is 340M;
    \item Training is by AWS p4d.24xlarge with 8 Nvidia Tesla A100 GPUs. PyTorch DistributedDataParallel is used in the trainer. Training data is randomly shuffled;
    \item All models performance are sensitive to Learning Rate in training; The exact value is algorithm dependent.
\end{itemize}

\begin{table}
 \center
  \caption{Hyperparameters of the three type of feature-crossing modules (MLP, MaskNet, Transformer) in DHEN.}
  \begin{tabular}{rc}
    \toprule
    
    \multirow{2}{*}[-0.2cm]{\begin{tabular}{r}MaskNet\end{tabular}}
    &\#MaskBlock=2, horizontal MaskBlock layout \\
    \cmidrule{2-2}
    &\#HiddenSize=256, Dropout=0.005 \\
    \midrule

    \multirow{3}{*}[-0.2cm]{\begin{tabular}{r}Transformer\end{tabular}}
    &\#Layer=2, \#Head=4 \\
    \cmidrule{2-2}
    &\#HiddenSize=256, \#Forward MLP Size=512 \\
    \cmidrule{2-2}
    &DropOut=0 \\
    \midrule
    
    MLP &Both MLP have the same output size 1024\\
    
    \bottomrule
  \end{tabular}
\label{table:hyperparameter}
\end{table}

\end{document}